\documentclass[twoside]{article}
\usepackage{amsfonts,amssymb,amsbsy,textcomp,marvosym,picins,amsmath,caption,threeparttable,amsthm,subfigure,float,lastpage,lscape}
\usepackage{eurosym,mathrsfs,fancyhdr,CJK,multicol,graphics,indentfirst,color,bm,upgreek,booktabs,graphicx,multirow,warpcol}
\usepackage{epstopdf}
\usepackage{graphicx}
\usepackage{multirow}
\usepackage{adjustbox}
\usepackage{wrapfig}
\usepackage{setspace}
\usepackage{algorithm}
\usepackage{algorithmic}
\usepackage{hyperref}
\def\footnoterule{\kern 1mm \hrule width 10cm \kern 2mm}

\allowdisplaybreaks
\catcode`@=11
\def\title#1{\vspace{3mm}\begin{flushleft}\vglue-.1cm\Large\bf\boldmath\protect\baselineskip=18pt plus.2pt minus.1pt #1
\end{flushleft}\vspace{1mm} }
\def\author#1{\begin{flushleft}\normalsize #1\end{flushleft}\vspace*{-4pt} \vspace{3mm}}
\def\address#1#2{\begin{flushleft}\vglue-.35cm${}^{#1}$\small\it #2\vglue-.35cm\end{flushleft}\vspace{-2mm}\par}
\def\jz#1#2{{$^{\footnotesize\textcircled{\tiny #1}}$\let\thefootnote\relax\footnotetext{\!\!$^{\footnotesize\textcircled{\tiny #1}}$#2}}}
\catcode`@=11
\def\section{\@startsection{section}{1}{\z@}%
 {-3ex \@plus -.3ex \@minus -.2ex}%
 {2.2ex \@plus.2ex}%
{\normalfont\normalsize\protect\baselineskip=14.5pt plus.2pt minus.2pt\bfseries}}
\def\subsection{\@startsection{subsection}{2}{\z@}%
 {-3ex\@plus -.2ex \@minus -.2ex}%
 {2ex \@plus.2ex}%
{\normalfont\normalsize\protect\baselineskip=12.5pt plus.2pt minus.2pt\bfseries}}
\def\subsubsection{\@startsection{subsubsection}{3}{\z@}%
 {-2.2ex\@plus -.21ex \@minus -.2ex}%
 {1.4ex \@plus.2ex}
{\normalfont\normalsize\protect\baselineskip=12pt plus.2pt minus.2pt\sl}}

\begin{document}
\begin{CJK*}{GBK}{song}
\thispagestyle{empty}
\vspace*{-13mm}
\noindent {\small }
\vspace*{2mm}

\title{Privacy-Enhanced Training-as-a-Service for On-Device Intelligence: Concept, Architectural Scheme, and Open Problems} 












\author{
Zhiyuan Wu$^{1,2}$,  
Sheng Sun$^{1}$,  
Yuwei Wang$^{1}$,  
Min Liu$^{1,3}$,  
Bo Gao$^{4}$,  
Tianliu He$^{1,2}$,  
Wen Wang$^{1,2}$,
}

\address{1}{Institute of Computing Technology, Chinese Academy of Sciences}
\address{2}{University of Chinese Academy of Sciences}
\address{3}{Zhongguancun Laboratory}
\address{4}{Beijing Jiaotong University}

\noindent
\textit{Contact Email: ywwang@ict.ac.cn}

\let\thefootnote\relax\footnotetext{{}\\[-4mm]\indent\ Regular Paper}

\noindent {\small\bf Abstract} \quad  {\small On-device intelligence (ODI) enables artificial intelligence (AI) applications to run on end devices, providing real-time and customized AI inference without relying on remote servers. However, training models for on-device deployment face significant challenges due to the decentralized and privacy-sensitive nature of users' data, along with end-side constraints related to network connectivity, computation efficiency, etc. Existing training paradigms, such as cloud-based training, federated learning, and transfer learning, fail to sufficiently address these practical constraints that are prevalent for devices. To overcome these challenges, we propose Privacy-Enhanced Training-as-a-Service (PTaaS), a novel service computing paradigm that provides privacy-friendly, customized AI model training for end devices. PTaaS outsources the core training process to remote and powerful cloud or edge servers, efficiently developing customized on-device models based on uploaded anonymous queries, enhancing data privacy while reducing the computation load on individual devices. We explore the threat model, definition, goals, and design principles of PTaaS, alongside emerging technologies that support the PTaaS paradigm. An architectural scheme for PTaaS is also presented, followed by a series of open problems that set the stage for future research directions in the field of PTaaS.}

\vspace*{3mm}


\vspace*{4mm}

\end{CJK*}
\baselineskip=18pt plus.2pt minus.2pt
\parskip=0pt plus.2pt minus0.2pt
\begin{multicols}{2}

\section{Introduction}
On-device intelligence (ODI) \cite{dhar2021survey,xu2024device} is an emerging technology that combines mobile computing and artificial intelligence (AI), empowering end-side on-device models to deliver real-time, customized intelligent inference without network connectivity. ODI shows considerable potential in the forthcoming Internet of Everything (IoE) era, as evidenced by applications such as intelligent medical diagnosis, AI-enhanced motion tracking \cite{iotj}, and behavior fingerprinting \cite{sanchez2021survey}. Leading technology corporations such as Google, Anthropic, and Xiaomi have recognized ODI as an essential technology in their flagship products, represented by Gemini-Nano \cite{team2023gemini}, Claude-3-Haiku \cite{claude2024}, and MiLM-1.3B \cite{xiaomi2023}.

Despite these advancements, ODI faces significant challenges due to the decentralized and privacy-sensitive nature of user data derived from the network termination \cite{murshed2021machine,hoofnagle2019european}. Hence, it is difficult for a single device to train a powerful AI model independently, especially with the massive end-side limitations related to network connectivity, computation capability, etc \cite{iotj}. 
To navigate these constraints and harness the full potential of ODI, various methods have been proposed to seek a balance between the need for powerful, customized AI training services and the practical constraints inherent in end devices.

\begin{figure*}[t]
	\centering
	\includegraphics[width=0.75\textwidth]{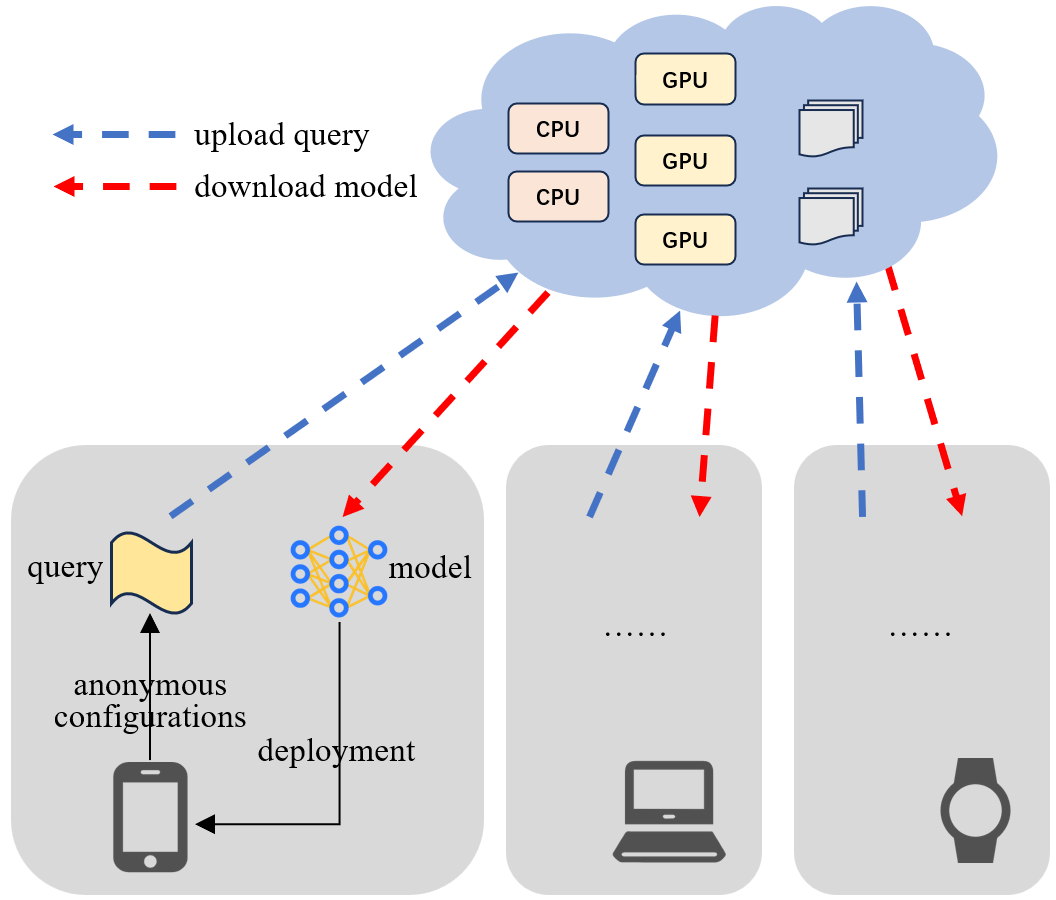}
	\caption{Schematic diagram of PTaaS.}
	\label{PTaaS-arch}
\end{figure*}

Cloud-based paradigms \cite{dhar2021survey,team2023gemini}, which involve uploading data from end devices to the cloud for centralized AI model training, offer intelligent inferences through offline deployment or application programming interfaces (APIs). These paradigms, however, raise concerns about user privacy \cite{dhar2021survey}, as devices are required to share their raw data with the cloud during the training process. 
Federated learning (FL) \cite{yang2019federated,wu2024fedcache,wu2023fedict} allows devices to collaboratively train AI models coordinated by a central server without the need for private data to leave devices. Whereas FL requires frequent network communication throughout the training process, which is not feasible for a mass of devices that may experience intermittent connectivity \cite{zhang2022federated}. 
Another approach, transfer learning (TL) \cite{patil2022poet,lin2022device,cai2020tinytl}, offers a solution of training generic base models in the cloud and then fine-tuning them on devices. Nevertheless, this process requires devices to participate in cumbersome training and inference built upon large-scale backbone models, which is impractical for devices with limited computation and energy resources. In summary, existing cloud-based paradigms fail to meet the privacy requirements of end devices. Although FL and TL can enhance model performance and data privacy, they encounter practical challenges related to network connectivity and computation efficiency. These issues motivate us to explore a novel paradigm that outsources on-device model training as a service to the remote cloud or edge servers in a privacy-enhanced manner.

In this paper, we propose the concept of Privacy-Enhanced Training-as-a-Service (PTaaS), which is a novel service computing paradigm for training AI models to be deployed on end devices, as shown in Fig. \ref{PTaaS-arch}. PTaaS offers privacy-enhanced, flexible, and efficient cloud or edge-based training services to address the challenges of data privacy, network connectivity, resource constraints, and customized requirements that are prevalent during model training procedures in ODI. The advantages of PTaaS are reflected in the following aspects:
\begin{itemize}
    \item 
    \textbf{Privacy enhanced.} PTaaS only requires devices to provide anonymous information related to local data as part of queries, eliminating the need for end devices to share local data with remote servers, hence enhancing user privacy.
    \item 
    \textbf{Centralized training.} PTaaS can fully utilize powerful computing resources and abundant open-source data owned by cloud or edge servers, training models to be deployed on devices with improved performance based on device-specific queries. At the same time, end devices do not need to afford the huge computation overhead during the training process, significantly reducing end-side computation and energy consumption.
    \item 
    \textbf{Simplicity and flexibility.} PTaaS migrates model training overhead from devices to the cloud, streamlining user operations to simply provide one-shot anonymous information and deploy downloaded trained models. As model training is decoupled from inference, devices can flexibly request model updates according to their customized demands, better adapting to dynamically changing application scenarios.
    \item 
    \textbf{Cost fairness and profit potential.} PTaaS service costs can be priced according to the consumed computing and data resources, ensuring fairness for diversified devices and enhancing their motivation to participate in AI model training. This pricing paradigm also enables service providers to earn reasonable profits, facilitating the wider adoption of PTaaS.
\end{itemize}

In the following sections, we provide a comprehensive elaboration on PTaaS from the perspectives of the threat model, definition, goals, and design principles. Besides, we analyze the emerging technologies that support PTaaS, including privacy computing, cloud-edge collaboration, transfer learning, and information retrieval. Furthermore, we provide the PTaaS hierarchy structure consisting of five layers: infrastructure, data, algorithms, services, and applications. Finally, we discuss a series of open problems faced by PTaaS in terms of improving privacy protection mechanisms, cloud-edge resource collaborative management, optimization of customized model training, designing pricing strategies, and standard specifications.

\color{black}
\section{Threat Model}
We establish a threat model that characterizes the security assumptions and potential vulnerabilities within the PTaaS service paradigm. The model encompasses the following key entities and their security properties:
\begin{itemize}
    \item \textbf{Remote servers.} The remote servers (both cloud and edge) are assumed to be honest but curious \cite{lindell2005secure}, meaning they faithfully execute the prescribed protocols but may attempt to infer sensitive information from the received data. While the servers are authorized to decrypt uploaded information for computation purposes, they are not trusted to access raw data generated on end devices.

    \item \textbf{Communication channels.} The communication channels between devices and servers are vulnerable to malicious third-party attackers who may attempt to eavesdrop on or tamper with transmitted data \cite{butun2019security}. These adversaries are assumed to have the capability to monitor and intercept all information passing through the network infrastructure.

    \item \textbf{End devices.} End devices are trusted to generate and maintain data locally while executing assigned tasks. However, sharing raw local data on devices could cause inadvertent leakage of sensitive information, which should be strictly prohibited \cite{hoofnagle2019european}.
\end{itemize}
This threat model guides the design of PTaaS security mechanisms, ensuring robust privacy protection while maintaining system functionality.
\color{black}

\section{Concept Elaboration}


\subsection{Definition of PTaaS}
PTaaS is a novel service computing paradigm that offers privacy-enhanced on-device model training as a service for ODI. It outsources the training of customized AI models to the remote cloud or edge servers, where the servers utilize their computing power and data resources to optimize models based on one-shot anonymous queries related to local data information provided by end devices. This service architecture separates model training and inference, reducing the training computation burden of devices while delivering high-performance and customized on-device models with enhanced data privacy.

\subsection{Goals of PTaaS}
The primary goal of PTaaS is to facilitate the efficient development of high-performance, customized on-device models with privacy enhancement while addressing the practical constraints faced by end devices, such as computation resources, network connectivity, etc. By outsourcing the training of on-device models to third-party cloud or edge servers, PTaaS aims to achieve the following objectives:
\begin{itemize}
\item
\textbf{Privacy protection.} PTaaS aims to protect user privacy by eliminating the need for end devices to share raw data with remote servers. Instead, devices are to provide only necessary anonymous information related to local data as part of queries, ensuring that sensitive user information remains private. 

\item
\textbf{Improved efficiency.} PTaaS aims to optimally exploit the sufficient computing power and the abundant data resources available in the cloud or the edge to train on-device models efficiently. This allows devices to benefit from high-performance and quickly available end-side models without extensive local resource consumption.

\item
\textbf{Flexibility and scalability.} PTaaS aims to accommodate the diverse and dynamic training requirements of end devices. It is intended to support training on-device models across various types, sizes, and user preferences while enabling periodic updates of model parameters to adapt to dynamically changing application scenarios. This ensures customized on-device models remain effective over time.

\item
\textbf{Accessibility and cost-effectiveness.} PTaaS aims to provide ubiquitous access to powerful AI training capabilities by providing a simplified interface and service process. In addition, it seeks to promote fair training by delivering on-demand services at fair prices built on transparent task scheduling and resource allocation. These enable users to participate equally in training tasks without understanding the complex details of the service process.
\end{itemize}

By focusing on the aforementioned goals, PTaaS seeks to address the limitations for ODI and provide a comprehensive solution for the efficient training of high-performance and customized on-device models with enhanced privacy.

\subsection{Design Principles of PTaaS}
The design of PTaaS should adhere to the following principles to ensure its effectiveness and practicality for ODI:
\begin{itemize}
\item 
\textbf{Privacy protection is paramount.} PTaaS should prioritize user privacy by ensuring all data circulation and processing procedures comply with related data protection regulations. Any information uploaded from devices to the cloud or edge servers should be strictly anonymous to prevent tracing back to individual private information. The principle of minimal data authorization should be followed, ensuring that only the information necessary for model training is sent to the cloud. Full user control over data-related information should be supported, including when to share, what to share, and the ability to withdraw shared information.

\item
\textbf{Resource optimization is the essence.} PTaaS should optimize the use of remote computing infrastructure through elastic resource scaling and rapid deployment. Training tasks should be dynamically assigned to the most suitable computing nodes based on intelligent scheduling algorithms customized for PTaaS, with various factors such as computing power, storage space, and network conditions sufficiently considered. PTaaS should also support heterogeneous computing, enabling collaboration between CPUs, GPUs, and TPUs to accommodate various types and structures of required models, thereby maximizing the efficiency of the training process.

\item
\textbf{Flexibility and scalability are required safeguards.} PTaaS should be designed as a highly flexible and scalable service architecture, providing APIs for users to configure customized training requirements and constraints, such as data resource utilization, computation requirements, budget constraints, and urgency levels. PTaaS platforms should continuously integrate new AI technologies and algorithms, support different scales of AI model training, and enable periodic updates and maintenance of on-device models. Moreover, PTaaS should be designed with ubiquitous accessibility in mind, supporting a wide range of device types as well as operating systems, and providing a universally adaptable deployment solution for end devices.

\item
\textbf{Cost fairness and profitability are key considerations.} PTaaS should ensure cost fairness and sustainable profitability in its business blueprint design. It should offer diverse cost options and enhance its market profitability by providing value-added services, such as model selection consulting and post-deployment technical support. Extensive partnerships can also be established to expand the functions and coverage of PTaaS services, creating a win-win situation for all kinds of service providers.
\end{itemize}


\begin{figure*}[t]
	\centering
	\includegraphics[width=0.55\textwidth]{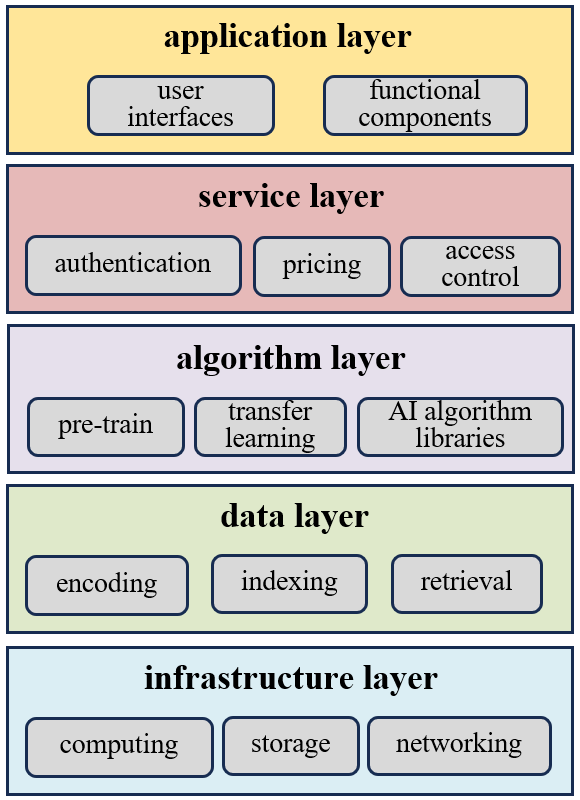}
	\caption{Overview of the five-layer hierarchy structure of PTaaS.}
	\label{PTaaS-hier}
\end{figure*}


\section{Architectural Scheme}
\subsection{Hierarchy Structure of PTaaS}
Drawing inspiration from the hierarchical architecture design of computer systems \cite{tanenbaum2016structured}, we propose the PTaaS hierarchy structure consisting of five layers: infrastructure layer, data layer, algorithm layer, service layer, and application layer, as shown in Fig. \ref{PTaaS-hier}. Each layer plays a distinct role in the design of PTaaS, and the layers collaborate to support the smooth operation of the PTaaS platform.
\begin{itemize}
\item 
\textbf{Infrastructure layer.} The infrastructure layer forms the foundation of PTaaS, providing the necessary physical resources such as computing, storage, and networking for the upper layers. The infrastructure layer ought to provide highly available, high-performance, and scalable computing as well as storage capabilities to meet the demands of training massive on-device models with dynamically changing working loads.

\item 
\textbf{Data layer.} The data layer manages and processes the remote data resources utilized by PTaaS. This layer includes data processing components such as data encoding, indexing, and retrieval, which work together to provide customized data resources for the above algorithm layer based on flexible user queries.

\item
\textbf{Algorithm layer.} The algorithm layer is the core of PTaaS and is primarily responsible for implementing model training algorithms. This layer includes unified AI algorithm libraries with diverse training acceleration mechanisms to handle different data and model characteristics. It also integrates transfer learning technology, leveraging pre-trained models on remote servers to improve the performance of models to be deployed on devices.

\item
\textbf{Service layer.} 
The service layer encapsulates the technical details of the infrastructure, data, and algorithm layers, providing an easy-to-use API for the application layer. This includes management functions such as authentication, pricing, and access control. The service layer receives parsed model training queries from the application layer, launches training tasks, and schedules them for execution in the algorithm layer according to request configurations.
    
\item 
\textbf{Application layer.} 
The application layer is the user-facing interface of PTaaS, offering user interaction components and extensive functionality for end devices. This layer includes various user interfaces, such as web pages, mobile apps, and command-line tools, alongside functional components represented by model selection, version control, and performance evaluation. The application layer allows users to initiate model training queries in a self-service manner and provides an intuitive visualization interface to display training progress, resource utilization, as well as model performance indicators, enabling users to monitor the training process and quality in real time.
\end{itemize}

As a whole, the five-layer PTaaS hierarchy structure is a bottom-up, hierarchical architecture with clearly defined roles for each layer. This layered architecture not only facilitates standardized design and development of PTaaS platforms, but also allows each layer to evolve and upgrade independently without affecting other layers. Thus, it empowers PTaaS platforms to respond flexibly to changes in emerging technologies and user requirements, supporting continuous innovation and maintenance.

\subsection{Emerging Technologies Related to PTaaS}
The realization of PTaaS relies heavily on several emerging technologies that ensure privacy, facilitate efficient training, and improve model performance. These technologies include privacy computing, cloud-edge collaboration, transfer learning, and information retrieval.
\begin{itemize}
    \item 
    \textbf{Privacy computing.} PTaaS platforms rely on advanced privacy computing technologies to safeguard the security of user data. Data desensitization methods, such as encryption and anonymization, can prevent identifiable information from being linked to uploaded queries. Sensitive hashing converts data into a commonly non-traceable form, allowing for secure operations such as sample retrieval on remote servers. 
    Moreover, differential privacy can be incorporated to increase the difficulty for service providers in inferring personal information from uploaded queries by introducing noise. In addition, the trusted execution environment can safeguard computation processes, blocking external attackers from prying into training information.
    \item
    \textbf{Cloud-edge collaboration.} 
    Cloud-edge collaboration is a crucial technique to accelerate the execution process of PTaaS. The cloud with abundant computing and storage resources undertakes the primary responsibilities of PTaaS training tasks, enabling the parallel optimization of extensive models. Complementing this, edge computing offers a strategic advantage by utilizing its proximate and low-latency resources to pre-process and manage training tasks. When the model training task is not heavy, devices can directly call the model training service from the accessed edge server, reducing both the service latency and the training load on the cloud.
    \item
    \textbf{Transfer learning.} Transfer learning boosts the effectiveness of PTaaS by making full use of extensive data resources and pre-trained models on the cloud or edge servers to enhance the performance of trained on-device models. Essentially, the servers train generic models on a vast amount of open-source data to capture general knowledge and patterns, helping to search for a strong starting point for on-device models to be trained. Models to be deployed on devices are then fine-tuned with queries-related data on servers, resulting in high-performing on-device models tailored to users' specific requirements.
    \item
    \textbf{Information retrieval.} 
    Efficient and privacy-enhanced retrieval is crucial for accessing valuable data within the PTaaS paradigm. PTaaS providers leverage queries uploaded from devices to retrieve the most relevant data samples on the cloud or edge servers for training customized models to be deployed on devices. Advanced information retrieval not only empowers PTaaS to quickly and accurately discover valuable data for model training but also supports various applications such as model recommendations and sensitive information filtering, further enhancing the functionality and usability of PTaaS platforms.
\end{itemize}

\begin{figure*}[t]
	\centering
	\includegraphics[width=1.00\textwidth]{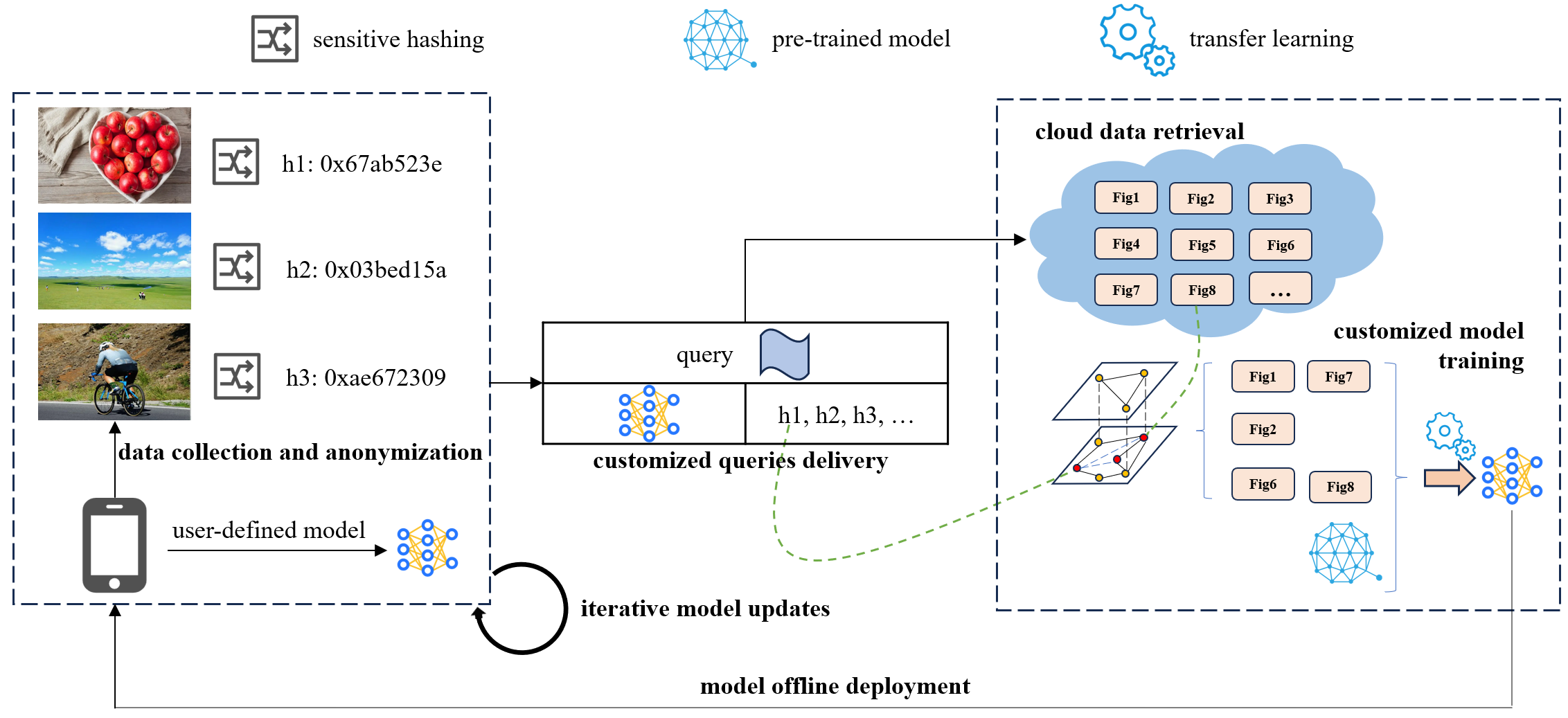}
	\caption{Framework of the instantiated PTaaS process based on cloud computing.}
	\label{PTaaS-process}
\end{figure*}

\begin{figure*}[t] 
    \centering
    \begin{minipage}{\textwidth}
\begin{algorithm}[H]
	\caption{Privacy-Enhanced Training-as-a-Service (PTaaS).}
	\label{alg1}
        procedure \textbf{DeviceExecute}()\\
        \hspace*{2em}//iterative model updates\\
        \hspace*{2em} \textbf{while} require\_model\_update is \textbf{true}: \\       
        \hspace*{4em} //data collection and anonymization\\
        \hspace*{4em} device\_data = collect\_data(end\_device)\\
        \hspace*{4em} device\_model = create\_model(end\_device)\\
        \hspace*{4em} anonymized\_data = hash\_function(device\_data)\\
        \hspace*{4em} differentially\_private\_anonymized\_data =  add\_differential\_privacy\_noise(anonymized\_data)\\
        \hspace*{4em} //customized queries delivery\\
        \hspace*{4em} query = encrypt(differentially\_private\_anonymized\_data, device\_model)\\       
        \hspace*{4em} transmit query to the cloud\\
        \hspace*{4em} \textbf{call} \textbf{CloudExecute}()\\
        \hspace*{4em} //model offline deployment\\
        \hspace*{4em} receive encrypted\_trained\_device\_model from the cloud\\
        \hspace*{4em} \textbf{if} check\_integrity(encrypted\_trained\_device\_model) is \textbf{true}:\\
        \hspace*{6em} trained\_device\_model = decrypt(encrypted\_trained\_device\_model)\\
        \hspace*{6em} deploy\_to\_inference\_engine(trained\_device\_model)\\
        end procedure\\
        
        procedure \textbf{CloudExecute}()\\
        \hspace*{2em} //customized queries delivery\\
        \hspace*{2em} receive query from end device\\
        \hspace*{2em} \textbf{if} verify\_query(query) is \textbf{true}:\\
        \hspace*{4em} differentially\_private\_anonymized\_data, device\_model = decrypt(encrypted\_query)\\
        \hspace*{4em} //cloud\_data\_retrieval\\
        \hspace*{4em} relevant\_data = similarity\_search(differentially\_private\_anonymized\_data, cloud\_dataset)\\
        \hspace*{4em} //customized model training\\
        \hspace*{4em} trained\_device\_model = transfer\_learning(device\_model, relevant\_data, pretrained\_model)\\
        \hspace*{4em} //model offline deployment\\
        \hspace*{4em} encrypted\_trained\_device\_model = encrypt(trained\_device\_model)\\
        \hspace*{4em} transmit encrypted\_trained\_device\_model to device\\
        end procedure
\end{algorithm}
    \end{minipage}
\end{figure*}

\subsection{PTaaS Process Instantiation}
To illustrate the on-demand training services offered by PTaaS, we provide a streamlined explanation of a simplified PTaaS process based on cloud computing, as shown in Fig. \ref{PTaaS-process} and \textcolor{black}{Algorithm \ref{alg1}}. This process demonstrates how PTaaS can be applied to image classification for smartphones.
\begin{itemize}
    \item
    \textbf{Data collection and anonymization.}
    Users collect image data of interest on their smartphones during their daily lives. Smartphones utilize sensitive hashing algorithms to anonymize the original images, generating hash vectors that preserve the images' characteristics while safeguarding private information. This process can be enhanced with differential privacy techniques, adding random noise to further protect user privacy. The smartphone also extracts image metadata, such as labels and shooting configurations, to provide additional context for model training.
    \item
    \textbf{Customized queries delivery.} The smartphones further encrypt the anonymous hash vectors, metadata, as well as users' customized image classification model (collectively referred to as queries), and then upload the queries to the cloud. The cloud verifies and decrypts the uploaded queries, refining users' model training requirements to ensure optimal trained results.
    \item
    \textbf{Cloud data retrieval.} The cloud employs privacy-enhanced similarity hash retrieval to identify highly relevant samples from its vast pool of data resources that match users' specific queries. This optimization of the dataset for training tasks takes advantage of the cloud's extensive data resources and sophisticated data management capabilities, paving the way for efficient and high-performance training of models to be deployed on devices.
    \item
    \textbf{Customized model training.} Utilizing the retrieved relevant data and pre-trained model resources on the cloud, PTaaS applies transfer learning to initialize and train customized models that are to be deployed on devices. The cloud's computation power, specialized training acceleration algorithms, along rich data repositories, significantly boost training efficiency while improving the performance of on-device models for the intended image classification task.
    \item
    \textbf{Model offline deployment.} The trained models are encrypted and transmitted back to users' smartphones, where they undergo decryption and an integrity check before being deployed to the local inference engine for practical inference tasks. This offline deployment ensures that the model can operate efficiently on devices without requiring cloud connectivity.
    \item 
    \textbf{Iterative model updates.} To accommodate dynamic changes in data distribution and evolving user requirements, PTaaS supports regular model updates. Users can upload new encrypted hash vectors and metadata to the cloud, utilizing which to train and deliver updated versions of the customized model. This iterative model update process ensures continuous maintenance of on-device models, guaranteeing models' effectiveness over time.
\end{itemize}
Depending on specific contexts, such as speech recognition or natural language processing, the process instantiation of PTaaS can be further optimized and adjusted for application-specific characteristics.

\section{Open Problems}
While PTaaS offers a promising solution for privacy-enhanced on-device AI model training, several open problems need to be addressed to ensure its practical implementation and widespread adoption. In this section, we discuss various open problems encountered by PTaaS, including improvement of privacy protection mechanisms, cloud-edge resource collaborative management, optimization of customized model training, designing pricing strategies, and standard specifications. We aim to provide insights for future research in the field of PTaaS.
\begin{itemize}
\item
\textbf{Improvement of privacy protection mechanisms.}
The instantiated PTaaS process utilizes advanced privacy computing technologies to protect user privacy. However, there is still the need for continuous improvement of these mechanisms in PTaaS to keep pace with evolving cyber threats and offer stronger guarantees against data breaches and misuse. At the same time, the dual demand for privacy protection and data utility in PTaaS presents a significant challenge, as excessive data desensitization may compromise the value of data and lead to poor training results. The question about how to adjust privacy-enhanced configurations based on the sensitivity of user local data and the specific requirements of training tasks is worth noting for future research in PTaaS.
\item
\textbf{Cloud-edge resource collaborative management.}
The current instantiated PTaaS process is confined to purely cloud-based service, while the evolving trend towards the shift in the computing paradigm from centralized to decentralized \cite{duan2022distributed} calls for the design of PTaaS to cloud-edge collaboration. However, the dynamic nature of cloud-edge architecture poses significant challenges in resource collaborative management over PTaaS. Efficiently leveraging the superior power of cloud computing and the proximity advantages of edge computing requires fine-grained task allocation and distributed resource scheduling algorithms for seamless collaboration between these two computing paradigms, considering the real-world computing operation of PTaaS such as decrypting and data retrieval to achieve the overall minimization of training latency and energy consumption.
\item
\textbf{Optimization of customized model training.} The instantiated PTaaS process offers customized AI model training services for devices with enhanced privacy. However, the wide range of user queries and the massive scale of server-side data present significant challenges to the PTaaS platform. Key issues include efficiently identifying the most valuable samples for customized model training within massive remote data, devising knowledge transfer strategies to enhance on-device model performance under provided data resources and balancing the full utilization of remote data resources against rising training costs.
\item 
\textbf{Designing pricing strategies.} Pricing strategies are vital in determining the sustainability, user adoption, market competitiveness, and profitability of PTaaS. The current PTaaS implementation lacks well-defined pricing strategies that consider the intricacy of training tasks, the utilization of computational and data resources, and the degree of model customization. Striking a balance between ensuring cost fairness across a diverse range of devices and securing reasonable profits for service providers is also essential, as it benefits both users and service providers alike.
\item
\textbf{Establishing standard specification.} The absence of unified standards in technical aspects, interfaces, and privacy policies greatly undermines the service incompatibility of PTaaS. Therefore, it is crucial to foster collaborative efforts toward establishing comprehensive standards for PTaaS, involving query exchange, service interfaces, and security protocols, involving academia, industry, and government sectors. Such cross-sector collaboration will establish a solid foundation for the scalable and sustainable development of PTaaS.
\end{itemize}

\color{black}
\section{Discussion}
\subsection{Security Mechanisms in PTaaS}
The security of PTaaS relies on three fundamental mechanisms: sensitive hashing, differential privacy, and encryption schemes, which work together to protect private on-device data during the service process.
For sensitive hashing, MinHash \cite{broder1997resemblance} and SimHash \cite{charikar2002similarity} can be applied for data anonymization and similarity-based retrieval, with MinHash ideal for set-based data, and SimHash excels in processing high-dimensional local data.
For differential privacy, the Laplace Mechanism \cite{dwork2006calibrating} is a standard choice for numerical anonymous data. In scenarios where the anonymous data follows a Gaussian distribution, the Gaussian Mechanism \cite{dwork2014algorithmic} is better suited due to its compatibility with the inherent statistical properties.
For encryption, AES \cite{daemen1999aes} provides a robust balance between security guarantees and computational efficiency, making it well-suited for resource-constrained devices.

\color{black}
\subsection{Feasibility Analysis of PTaaS}
The feasibility of PTaaS can be evaluated from two key perspectives: technological support and practical applications. From a technological perspective, PTaaS leverages several well-established technologies, including cloud computing, privacy computing, and transfer learning, which collectively provide a robust foundation for its implementation. The proposed five-layer hierarchical architecture further facilitates the modular development and scalability of PTaaS platforms, ensuring flexibility and adaptability to evolving technological trends.
From a practical perspective, PTaaS addresses critical real-world challenges in ODI deployment. Its ability to deliver privacy-enhanced, customized model training while offloading computational burdens from resource-constrained devices makes it particularly suitable for mobile scenarios with limited computational power, energy, or storage. This aligns with the growing demand for on-device AI applications across diverse environments. Moreover, the simplified one-shot query mechanism significantly lowers the barrier to entry for users, offering a more accessible alternative to existing approaches, such as federated learning and traditional transfer learning, which often require extensive communication or local computation. These advantages make PTaaS a promising paradigm for the widespread adoption of ODI-enabled applications.

\subsection{Privacy Level of PTaaS}
While PTaaS enhances device-side privacy in outsourcing the core training process, it is important to clarify the level of privacy protection it provides. By requiring devices to share only anonymous information derived from local data, PTaaS avoids the transmission of raw user data over the network, thereby delivering a meaningful improvement in privacy protection. Despite these superiorities, PTaaS enhances privacy rather than preserves it completely. Since service providers must decrypt and process certain anonymized information to train models, there remains a theoretical risk of privacy breaches. For instance, adversaries could potentially reconstruct private information by optimizing for hash similarity. As such, while PTaaS significantly improves data privacy compared to centralized training paradigms, it is not impervious to all privacy risks.

\color{black}

\section{Conclusions and Future Works}
In this paper, we propose Privacy-Enhanced Training-as-a-Service (PTaaS), a novel service computing paradigm for on-device intelligence (ODI). PTaaS aims to address the challenges faced by on-device model training with enhanced privacy by outsourcing the training of end-side artificial intelligence (AI) models to cloud or edge-based service providers, with only anonymous queries shared with remote servers. PTaaS enables the development of high-performance, customized on-device AI models with enhanced data privacy while alleviating the computation, storage, and energy constraints of end devices. Future works will focus on enhancing privacy protection mechanisms, optimizing cloud-edge resource collaborative management, improving customized model training, designing novel pricing strategies, and establishing standard specifications, so as to ensure the sustainable development of PTaaS.

 \bibliographystyle{unsrt}

\label{last-page}
\end{multicols}
\label{last-page}
\end{document}